\documentclass[letterpaper, 10 pt, conference]{ieeeconf}
\usepackage{amsmath}
\usepackage{amssymb}
\usepackage{mathtools}

\usepackage{tabularx}
\usepackage{graphicx}
\usepackage{bm}
\usepackage{color}
\usepackage{float}
\usepackage{units}
\usepackage{url}
\usepackage{hyperref}
\usepackage{balance}
\usepackage{amsmath}

\makeatletter 
\def\endfigure{\end@float} 
\def\endtable{\end@float}
\makeatother

\usepackage{subcaption}
\usepackage{caption}

\renewcommand{\unit}[1]{{\rm #1} }

\newcommand{\rev}[1]{\textcolor{black}{#1}}

\IEEEoverridecommandlockouts

\begin{document} 

% Paper title (needs to change)
\title{\Large \bf 			
Kinodynamics-based Pose Optimization for Humanoid Loco-manipulation
}
% Kinodynamics-based Pose Optimization and MPC with kinodynamics model 
% Kinodyanmics Pose Optimization and Loco-manipulation MPC for Enhanced object-pushing on Humanoid Robots
% Pushing Heavy Objects via Pose-enhanced MPC on Humanoid Robots
% Optimized Pushing on humanoid robots via loco-manipulation MPC
% Pose-assisted MPC for Pushing Heavy Objects on Humanoid Robots
	
%\newtheorem{assumption}{Assumption}

\author{Junheng Li and Quan Nguyen\thanks{Junheng Li and Quan Nguyen are with the Department of Aerospace and Mechanical Engineering, University of Southern California, Los Angeles, CA 90089.
email:{\tt\small junhengl@usc.edu, quann@usc.edu}}%
}%													
	
% make the title area
\maketitle
%Pins
%Junheng Li 150116
% Quan 203386

\begin{abstract}

% This paper presents a novel approach for controlling humanoid robots pushing heavy objects using kinodynamics-based pose optimization and loco-manipulation MPC. The proposed pose optimization plans the optimal pushing pose for the robot while accounting for the unified object-robot dynamics model in steady state, robot kinematic constraints, and object parameters. 
% The approach is combined with loco-manipulation MPC to track the optimal pose.
% Coordinating pushing reaction forces and ground reaction forces, the MPC allows accurate tracking in manipulation while maintaining stable locomotion. 
% In numerical validation, the framework enables the humanoid robot to effectively push objects with a variety of parameter setups. The pose optimization generates different pushing poses for each setup and can be efficiently solved as a nonlinear programming (NLP) problem, averaging 250 ms. The proposed control scheme enables the humanoid robot to push an object with a mass of up to 20 kg (118$\%$ of the robot's mass). Additionally, \rev{the MPC can recover the system when a 120 N lateral force disturbance is applied to the object for 0.3 s.}

This paper presents a novel approach for controlling humanoid robots to push heavy objects. The approach combines kinodynamics-based pose optimization and loco-manipulation model predictive control (MPC). The proposed pose optimization considers the object-robot dynamics model, robot kinematic constraints, and object parameters to plan the optimal pushing pose for the robot. The loco-manipulation MPC is used to track the optimal pose by coordinating pushing and ground reaction forces, ensuring accurate manipulation and stable locomotion. Numerical validation demonstrates the effectiveness of the framework, enabling the humanoid robot to push objects with various parameter setups. The pose optimization can be solved as a nonlinear programming (NLP) problem within an average of 250 ms. The proposed control scheme allows the humanoid robot to push objects weighing up to 20 kg (118$\%$ of the robot's mass). Additionally, it can recover the system from a 120 N  lateral force disturbance applied for 0.3 s.

\end{abstract}

% Introduction and literature review

\section{Introduction}
\label{sec:Introduction}

%%%%%%%% Motivation %%%%%%%
% Humanoid robots have gained significant importance in recent years due to the increasing demand for more versatile and autonomous robots that can perform tasks in real-world environments. The development of humanoid robots has been driven by advancements in technology, such as computing power, sensors, and actuators. These advancements have made it possible to implement intricate controllers on robots to perform tasks that were previously only possible for humans, such as very dynamic walking, running, and manipulating objects \cite{kuindersma2016optimization, paredes2022resolved, QDH, kim2020dynamic, abi2019torque}. In this work, we intend to further investigate the potential of dynamic humanoid loco-manipulation. By  leveraging the whole-body poses on humanoid robots and coordinating pushing and locomotion in control, we can allow effective loco-manipulation control of large and heavy objects.

Humanoid robots have gained prominence for their versatility and autonomy in real-world environments. Technological advancements in computing power, sensors, and actuators have propelled the development of these robots. This has enabled the implementation of complex controllers, empowering them to perform dynamic tasks like walking, running, and object manipulation \cite{kuindersma2016optimization, paredes2022resolved, QDH, kim2020dynamic, abi2019torque}. In this study, we aim to explore the capabilities of dynamic humanoid loco-manipulation. By leveraging whole-body poses and coordinating pushing and locomotion, humanoid robots can achieve effective control of large and heavy objects.

% Legged robots have shown the potential to achieve various manipulation strategies, including the use of robotic arms on quadruped robots \cite{bellicoso2019alma,rehman2016towards} and highly flexible hands on humanoid robots \cite{liu2021soft,ott2006humanoid}. However, the grasping manipulation setups in \cite{liu2021soft,ott2006humanoid} are not suitable for heavy and large objects. 
% While humans push heavy objects, we tend to adjust our body pose based on different object dimensions, masses, and terrain roughness. Inspired by this, we propose a pose optimization framework to optimize for humanoid pushing poses to adapt to the object parameters and setups.

Legged robots have demonstrated their potential for various manipulation strategies, such as using robotic arms on quadruped robots \cite{bellicoso2019alma,rehman2016towards} and flexible hands on humanoid robots \cite{liu2021soft,ott2006humanoid}. However, the grasping setups described in \cite{liu2021soft,ott2006humanoid} are not suitable for handling heavy and large objects. In contrast, humans adjust their body pose when pushing heavy objects, considering factors such as object dimensions, mass, and terrain roughness. Motivated by this human behavior, we propose a pose optimization framework for humanoid robots to optimize their pushing poses based on object parameters and setups.

%%%%%%%%%%%%%%%%%%%%%%%%%%%%%%%%%

%%%%%%%%%%%% MPC %%%%%%%%%%%%
% Recently, MPC-based approaches for legged robot locomotion control have become popular. The MIT Cheetah 3 \cite{di2018dynamic} and Mini Cheetah \cite{katz2019mini, kim2019highly} showcase dynamic locomotion capabilities using force-based MPC. Bipedal/humanoid robots have also achieved impressive results in balancing after acrobatic motions \cite{chignoli2021humanoid} and dynamically traversing uneven terrain \cite{li2023dynamic} using MPC-based control schemes.  
% The authors' previous work \cite{li2021force,li2022multi} developed MPC-based bipedal locomotion control and humanoid loco-manipulation control.
% In \cite{li2022multi}, MPC simplifies the object dynamics as external forces applied to the robot in loco-manipulation. 
% Authors in \cite{sleiman2021unified} proposed a unified whole-body loco-manipulation planner via non-linear MPC on quadruped robots with a robotic arm.
% However, these works do not focus explicitly on optimizing the robot's whole-body poses with the consideration of object parameters.
% In our work, we aim to leverage non-prehensile loco-manipulation to move large and heavy objects. On humanoid robots, both hands can be used for pushing, which means the object can be controlled more effectively compared to a single contact point for manipulation in \cite{rigo2022contact} and \cite{sleiman2021unified}. Hence, we  leverage humanoid robots' double-hand pushing contacts and task-oriented whole-body poses to achieve dynamic loco-manipulation. 

MPC-based approaches have gained popularity for legged robot locomotion control. Notable examples include the MIT Cheetah 3 \cite{di2018dynamic} and Mini Cheetah \cite{katz2019mini, kim2019highly}, which utilize force-based MPC for dynamic locomotion. Bipedal/humanoid robots have also demonstrated impressive capabilities in balancing after acrobatic motions \cite{chignoli2021humanoid} and traversing uneven terrain \cite{li2023dynamic} using MPC-based control schemes. The authors' previous work \cite{li2021force,li2022multi} focused on MPC-based bipedal locomotion control and humanoid loco-manipulation control. In \cite{li2022multi}, MPC simplifies object dynamics as external forces applied to the robot during loco-manipulation. Another study \cite{sleiman2021unified} proposed a unified whole-body loco-manipulation planner using nonlinear MPC on quadruped robots with a robotic arm. However, these works do not explicitly address optimizing the robot's whole-body poses while considering object parameters. In this study, we aim to leverage non-prehensile loco-manipulation to move large and heavy objects. By using both hands for pushing on humanoid robots, we can achieve more effective control compared to single contact point manipulation as seen in \cite{sleiman2021unified} and \cite{rigo2022contact}. Therefore, we utilize the double-hand pushing contacts and task-oriented whole-body poses of humanoid robots to achieve dynamic loco-manipulation.

%%%%%%% title figure %%%%%%%%%
\begin{figure}[!t]
\vspace{0cm}
\captionsetup[subfigure]{justification=centering}
     \centering
     \begin{subfigure}[b]{0.13\textwidth}
         \centering
         \includegraphics[clip, trim=0cm 7cm 24.3cm 3cm, width=\columnwidth]{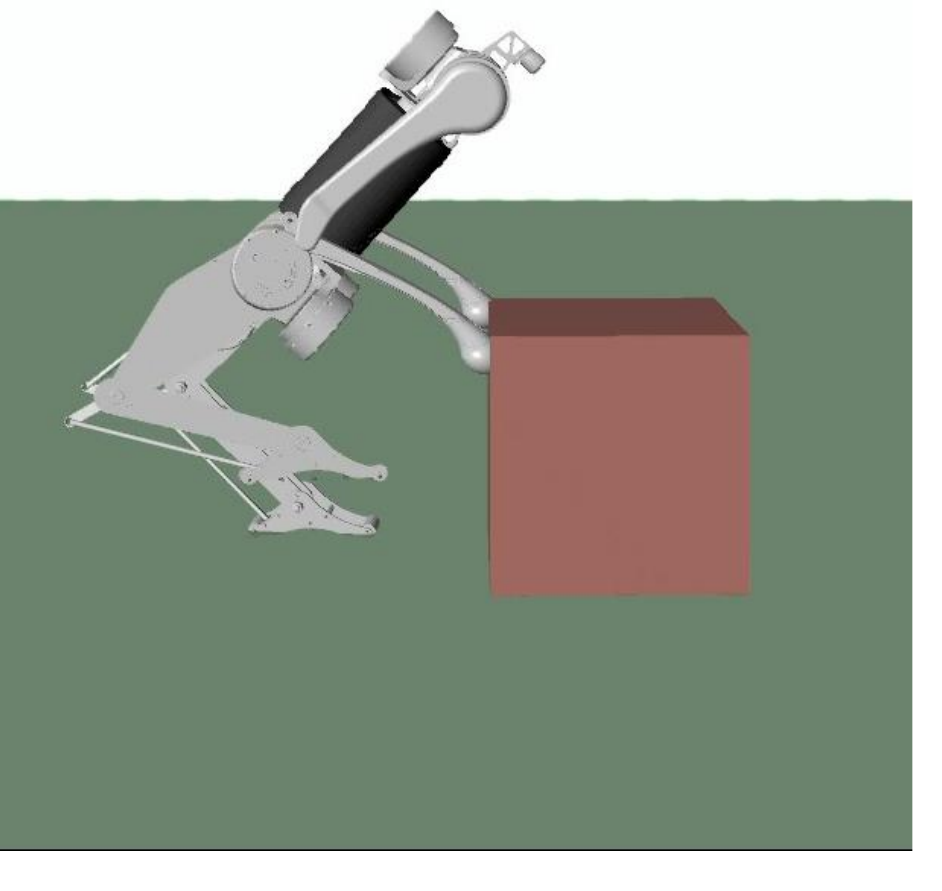}
         \caption{5 kg, 0.3 m cube, \\ $\mu_g$ = 0.7}
         \label{fig:po1}
     \end{subfigure}
     \hfill
     \begin{subfigure}[b]{0.15\textwidth}
         \centering
         \includegraphics[clip, trim=10cm 7cm 13.3cm 3cm, width=\columnwidth]{Figures_new/title.pdf}
         % \begin{center}
            \caption{10 kg, 0.5 m cube, \\ \quad $\mu_g$ = 0.4}
         % \end{center}
         \label{fig:po2}
     \end{subfigure}
     \hfill
     \begin{subfigure}[b]{0.18\textwidth}
         \centering
         \includegraphics[clip, trim=20.9cm 7cm 0cm 3cm, width=\columnwidth]{Figures_new/title.pdf}
         \caption{15 kg, 0.7 m cube, \\ $\mu_g$ = 0.5}
         \label{fig:po3}
     \end{subfigure}
        \caption{{\bfseries Humanoid Robot Pushing Various Objects with Optimized Poses.} Simulation video: \url{https://youtu.be/iunrSQjQx4M}}
        \label{fig:title}
        \vspace{-0.3cm}
\end{figure}
%%%%%%%%%%%%%%%%%%%%%%%%%%%%

%%%%%%% planning, TO, PO %%%%%%%%%%%%%%
% Trajectory optimization is a popular motion planning technique (e.g., \cite{nguyen2019optimized, chignoli2021humanoid, li2023dynamic}). This technique optimizes the robot's body trajectory and foot positions over a period of time and then tracks optimal trajectories using real-time feedback control. Our approach, kinodynamics-based pose optimization, also acts as a planner for generating optimal pushing poses. The kinodynamics model in our work is defined as the object-robot simplified dynamics models with robot kinematic constraints. Instead of optimizing the robot's full trajectory, pose optimization solves only one optimal pose for the entire pushing action to generalize the pushing motion and reduce computational cost.
% The authors' previous work \cite{li2022balancing} introduced pose optimization as a means of a kinematic-based planner to allow wheel-legged quadrupedal robots to utilize their whole-body motion to overcome high obstacles. However, in this work, the kinematic-only model does not account for the effects of ground friction and object mass, which are essential aspects in non-prehensile loco-manipulation. Hence, it is important to consider the dynamics model of the system as well. In different object setups, the optimal pushing pose is task-oriented and not generalizable. Therefore, pose optimization is used to solve the optimal pose that is most suitable for the setup.   
Trajectory optimization is a widely used motion planning technique \cite{nguyen2019optimized, chignoli2021humanoid, li2023dynamic}. It involves optimizing the robot's body trajectory and foot positions over a specified time period and then tracking these optimal trajectories using real-time feedback control. In our approach, we propose kinodynamics-based pose optimization as a planner for generating optimal pushing poses. Our kinodynamics model incorporates simplified object-robot dynamics with robot kinematic constraints. Instead of optimizing the robot's complete trajectory, pose optimization aims to find a single optimal pose for the entire pushing action, allowing for generalized pushing motions and reducing computational complexity. In a previous work \cite{li2022balancing}, pose optimization was introduced as a kinematic-based planner for wheel-legged quadrupedal robots to overcome high obstacles using whole-body motion. However, that work relied solely on kinematics and did not account for ground friction and object mass, which are crucial factors in non-prehensile loco-manipulation. Moreover, the optimal pushing pose is task-oriented and specific to each object setup, rather than being generalizable. Hence, in this work, kinodynamics-based pose optimization is used to determine the most suitable optimal pose for each setup.

\begin{figure}[!t]
\vspace{0.2cm}
		\center		\includegraphics[clip, trim=3cm 3cm 4cm 0.3cm, width=\columnwidth]{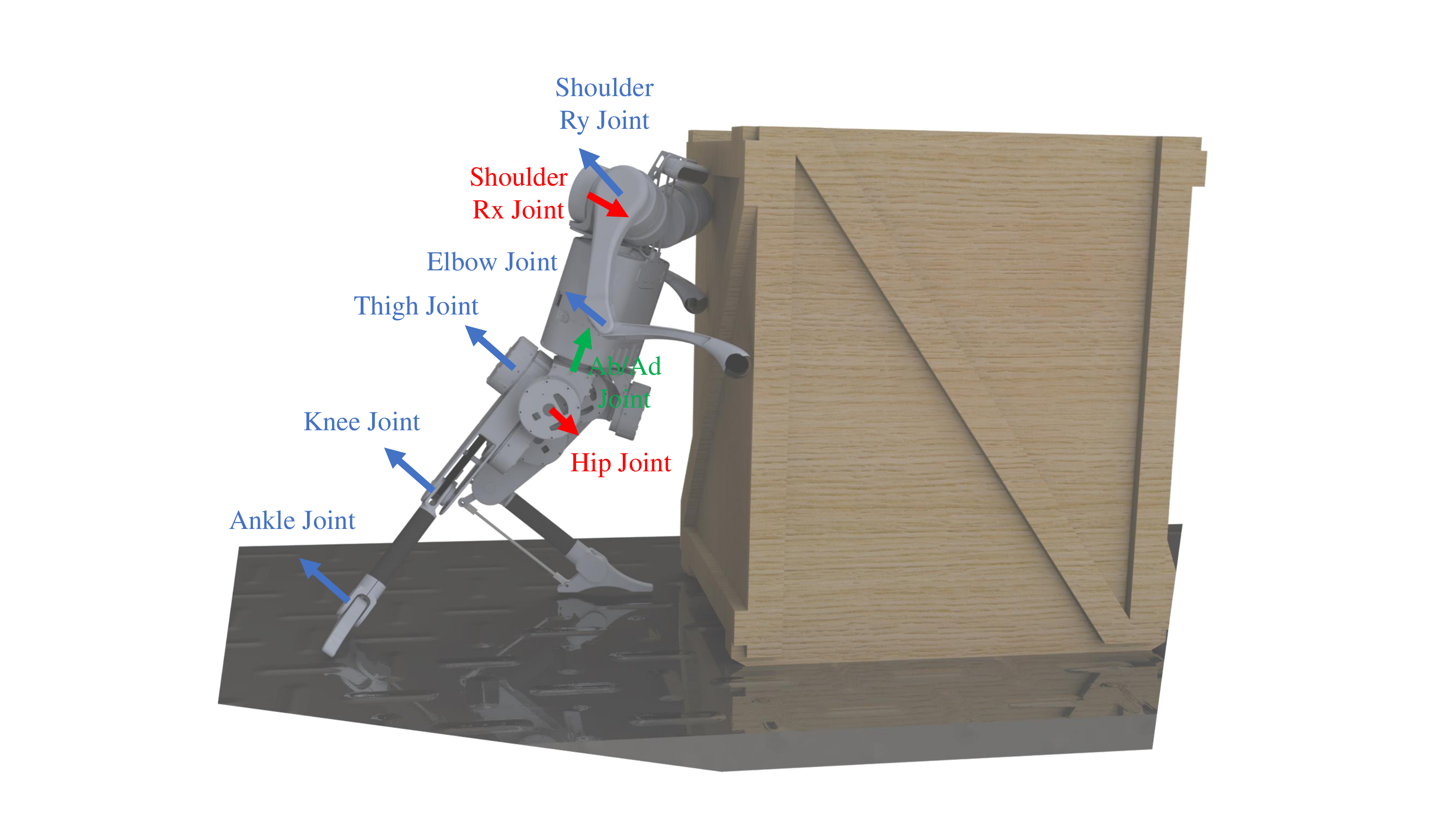}
		\caption{{\bfseries Humanoid Robot Model and Joint Definitions} \rev{(Pose in this render does not imply the resulting pose from the proposed approach)}}
		\label{fig:Robot}
		\vspace{-0.2cm}
\end{figure}

In \cite{murooka2021humanoid, ferrari2017humanoid}, non-prehensile loco-manipulation on humanoid robots was studied, employing techniques like RRT for planning object, body, or footstep paths throughout the entire trajectory. In another study \cite{murooka2015whole}, whole-body poses were used to apply a specific amount of pushing force during humanoid pushing. The desired poses and contact locations were determined based on the desired force using a randomization-based algorithm, and impedance control was employed to realize the desired behavior on humanoid robots. In our approach, we utilize pose optimization to generate whole-body poses while considering object parameters. The optimal pushing pose is obtained by solving an NLP problem that incorporates object-robot kinodynamics constraints. Unlike previous methods, we focus solely on generating a single pushing pose for the entire task rather than solving for object or robot trajectories within the pose optimization process. Additionally, we combine pose optimization with an online loco-manipulation MPC to track the optimal pose and coordinate whole-body manipulation and locomotion. Our MPC framework also allows for real-time responses to external disturbances.

%%%%%%%%%%%%%%%%% Main Contributions %%%%%%%%%%%%%%%%%%
% \subsection{Main Contributions}
% \label{sec:MainContribution}
The main contributions of the paper are as follows:
\begin{itemize}
    \item We developed a novel kinodynamic-based pose optimization framework to solve for optimal pushing poses on humanoid robots with the consideration of object-robot steady-state dynamics, kinematics, and physical properties. 

    \item Pose optimization generates unique poses that correspond to different setups, such as object size, object mass, and friction coefficients. The NLP problem can be efficiently solved with an average solving time of 250 $\unit {ms}$.
    
    \item We proposed a loco-manipulation MPC with a unified object-robot dynamics model, which optimizes pushing and ground reaction forces for coordinated whole-body loco-manipulation while tracking the optimal pose.
    
    % \item The non-prehensile loco-manipulation MPC allows the loss of contact and pushing force for one hand during pushing for object turning and pivoting.  

    \item The proposed framework can allow successful loco-manipulation control with a wide range of object parameter setups, including object side length in the range of 0.3 $\unit{m}$ to 1 $\unit{m}$, and object mass up to 20 $\unit{kg}$.

    \item The proposed control framework can recover deviations from desired states after introducing a 120 $\unit{N}$ lateral force for 0.3 $\unit{s}$ to the object, 

\end{itemize}
% \subsection{Overview}
% \label{sec:overview}

The rest of the paper is organized as follows. Section. \ref{sec:robotModel} presents the overview of the control system architecture. Section.\ref{sec:approach} introduces the details of the robot model, unified robot-object dynamics, pose optimization framework, and the loco-manipulation MPC. Some simulation result highlights and comparisons are presented in Section. \ref{sec:Results}.

% Robot Model and Simulation
\section{System Overview}

\label{sec:robotModel}
% \subsection{Robot Model}
% % \subsection{Robot Model}
% % \label{subsec:robotModel}
% In this section, we introduce the humanoid robot model employed in this work. The design, shown in Figure \ref{fig:Robot}, is followed from our previous work \cite{li2022multi} and features a compact humanoid robot with 5-DoF legs and 3-DoF arms.
% %Joint configurations are presented in Figure. \ref{fig:design}. 
% All joints are actuated by Unitree A1 torque-controlled motors with a maximum torque output of 33.5 $\unit{Nm}$ and maximum joint speed output of 21.0 $\unit{rad/s}$, except for the knee joints, which have a maximum torque output of 67 $\unit{Nm}$ due to halved gear ratio.

% We design the ankle joint to be actuated as well in order to allow the robot to balance with double-leg support with line foot, which is a design choice seen on Cassie \cite{gong2019feedback} and MIT humanoid robot \cite{chignoli2021humanoid}. The joint actuators are placed on the upper part of the thigh links on the legs and around the shoulders on the arms, for mass concentration and to minimize leg dynamics during locomotion. This is an important assumption in our simplified dynamics model that assumes negligible leg/arm mass. The humanoid robot has an overall mass of 17 $\unit{kg}$.

%%%%%%%%%%%%%%%%%%%%%%%%%%%%%%%%%%%%%%%%%%%%%%%%%%%%%%%%%%%%%%
\begin{figure}[!t]
\vspace{0.2cm}
		\center
		\includegraphics[clip, trim=2cm 3cm 4.5cm 1cm, width=\columnwidth]{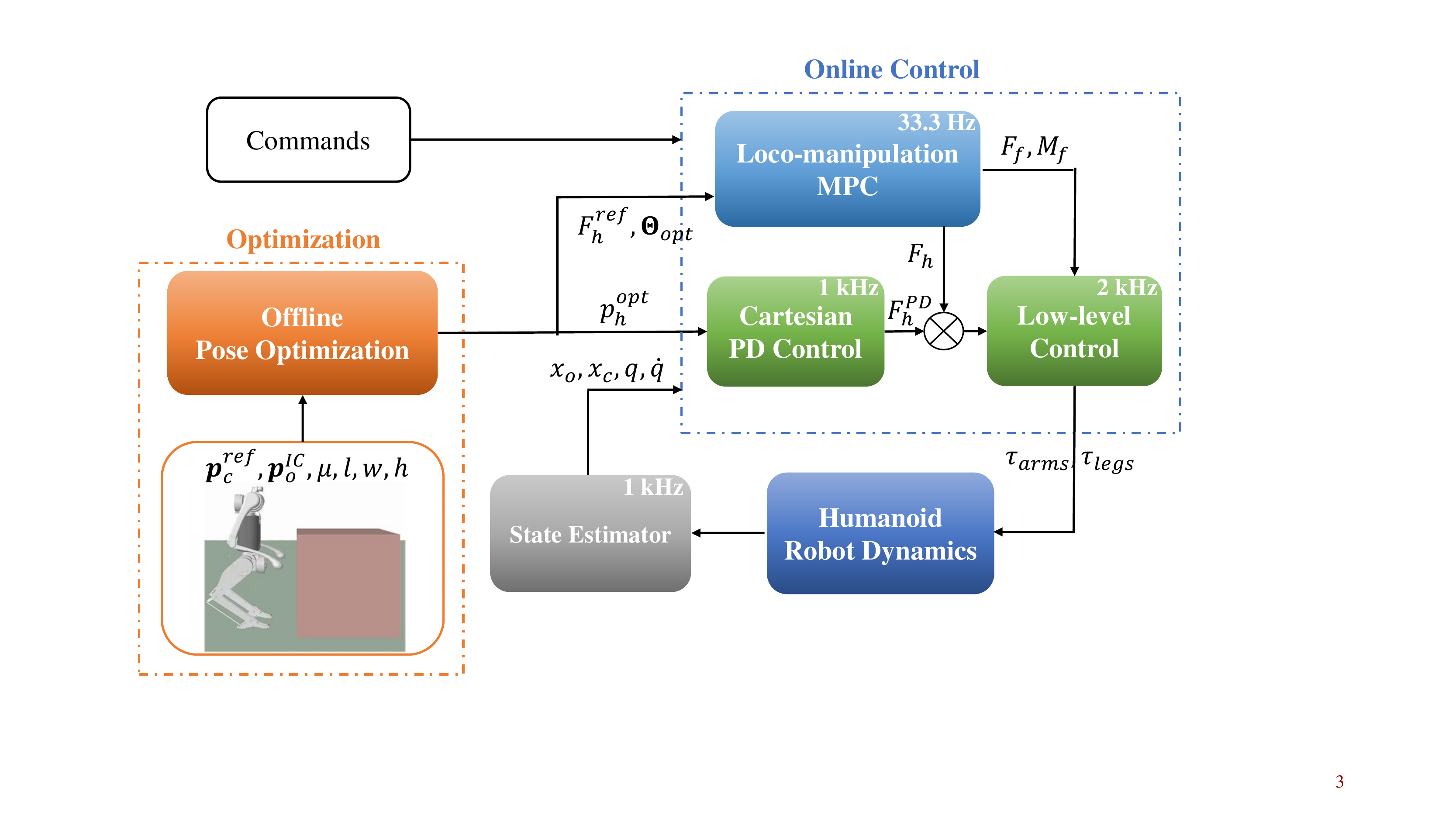}
		\caption{{\bfseries \rev{System Architecture}}}
		\label{fig:controlArchi}
		\vspace{-0.2cm}
\end{figure}

% \subsection{System Overview}
% \label{sec:sysoverview}

% In this section, we present the control system architecture of the proposed approaches, illustrated in Figure. \ref{fig:controlArchi}.
% Given the information of the object's initial states and physical properties, we first use the kinodynamics-based pose optimization to generate an optimal pushing pose based on the object dimension, mass, and friction coefficients. By using the kinodynamics model, under different object parameter setups, pose optimization generates different poses which ensure the robot is under a large stability region and hand contact locations are optimal for pushing action. The optimal pose and pushing reaction forces are then used as references for loco-manipulation MPC to track. The MPC leverages the unified dynamics of the object-robot system to coordinate the manipulation and locomotion tasks with whole-body pushing motion.

In this section, we introduce the control system architecture of the proposed approaches, depicted in Figure. \ref{fig:controlArchi}. Firstly, using the kinodynamics-based pose optimization, we generate an optimal pushing pose considering the object's initial states and physical properties, including dimension, mass, and friction coefficients. The kinodynamics model enables pose optimization to produce different poses for various object parameter setups, ensuring a large stability region and optimal hand contact locations for effective pushing action. Then, the obtained optimal pose and pushing reaction forces serve as references for the loco-manipulation MPC. The MPC leverages the unified dynamics of the object-robot system to coordinate the manipulation and locomotion tasks, incorporating whole-body pushing motion.

The kinodynamics-based pose optimization serves as a planner for solving the humanoid pushing pose.
The information used in pose optimization includes the object's initial CoM position $\bm p_o^{IC} \in \mathbb R^{3}$, the robot reference CoM position when initializing the pushing task $\bm p_c^{ref} \in \mathbb R^{3}$, friction coefficients at different contact locations, and object dimensions, where $l,w,$ and $h$ are the length, and width, and height of the rectangular object. 
% The optimal pushing pose is solved in terms of optimal body Euler angles $\bm \Theta^{opt}$, hand locations $\prescript{}{\mathcal{B}} {\bm p_{h,n}^{opt}} \in \mathbb R^{3}$ for $n^{th}$ hand, and foot locations $\prescript{}{\mathcal{B}}{\bm p_{f,m}^{opt} \in \mathbb R^{3}}$ for $m^{th}$ foot,  in the robot body frame, and $n = 1,2; \: m = 1,2$. 
The optimal pushing pose is solved in terms of optimal body Euler angles $\bm \Theta^{opt}$ and robot joint angles $\bm q  \in \mathbb R^{16} $.
Pose optimization also generates a reference pushing reaction force for MPC $\bm F_h^{ref} \in \mathbb R^{3}$. 

The loco-manipulation MPC serves as an online feedback controller to control the arms and stance leg for manipulating the object while performing stable locomotion. It tracks the optimal Euler angles and optimizes pushing reaction forces $\bm F_{h,n} \in \mathbb R^{3}$ and ground reaction forces and moments $\bm F_{f,m} \in \mathbb R^{3}, \bm M_{f,m} \in \mathbb R^{2}$.

A Cartesian PD controller is used in parallel with MPC for controlling the swing leg and hand locations. 
To compute arm and leg joint torques commands $\bm \tau_{arm,n} \in \mathbb R^{3}, \: \bm \tau_{leg,m} \in \mathbb R^{5}$, in low-level control, we use whole-body contact Jacobians $\bm J_c$ to map forces and moments to joint torques.

There are several state feedback information we acquire from the robot.
The robot state feedback $\bm x_c$ include body Euler angles (roll, pitch, and yaw) ${\bm \Theta_c = [\phi_c,\:\theta_c,\:\psi_c]}^\intercal$, CoM location $\prescript{}{\mathcal{}}{\bm p_c} \in \mathbb R^{3}$, velocity of body CoM  $\prescript{}{\mathcal{}}{\dot{\bm p}_c} \in \mathbb R^{3}$, and angular velocity $\bm \omega_c \in \mathbb R^{3}$.
The object state feedback $\bm x_o$ include object Euler angles ${\bm \Theta_o = [\phi_o,\:\theta_o,\:\psi_o]}^\intercal$, object CoM location $\prescript{}{\mathcal{}}{\bm p_o} \in \mathbb R^{3}$, velocity of object CoM  $\prescript{}{\mathcal{}}{\dot{\bm p}_o} \in \mathbb R^{3}$, and angular velocity $\bm \omega_o \in \mathbb R^{3}$.
Joint feedback  includes the joint positions $\bm q$ and velocities $\dot{\bm q} $ of the humanoid robot.

% Main theory 
\section{Proposed Approach}
\label{sec:approach}

In this section, we introduce the proposed approach established in this work, including the details of dynamics models of the loco-manipulation system, kinodynamics-based pose optimization, and loco-manipulation MPC.

\subsection{Robot Dynamics Models}
\label{subsec:CSMPC}

The humanoid robot design, shown in Figure \ref{fig:Robot}, is followed from our previous work \cite{li2022multi} and features a compact humanoid robot with 5-DoF legs and 3-DoF arms. The robot has an overall mass of 17 $\unit{kg}$.
%Joint configurations are presented in Figure. \ref{fig:design}. 
All joints are actuated by Unitree A1 torque-controlled motors with a maximum torque output of 33.5 $\unit{Nm}$ and maximum joint speed output of 21.0 $\unit{rad/s}$, except for the knee joints, which have a maximum torque output of 67 $\unit{Nm}$ due to halved gear ratio.

\rev{Firstly, we present the joint-space full dynamics equation of the humanoid robot. The joint-space generalized states ${\mathbf{q}} \in \mathbb{R}^{22}$ include $\bm p_c, \mathbf \Theta_c $, and $\bm q$.}
\begin{align}
\label{eq:fullDynamics}
    \rev{  \mathbf{H}(\mathbf{q})\ddot{\mathbf{q}} + \mathbf{C}(\mathbf{q}, \dot{\mathbf{q}}) = \mathbf{\Gamma} + \bm{J}_i(\mathbf{q})^\intercal \bm{\lambda}_i }
\end{align}

\rev{where $\mathbf{H} \in \mathbb{R}^{22 \times 22}$ is the mass-inertia matrix and $\mathbf{C} \in \mathbb{R}^{22}$ represents the joint-space bias force. $\mathbf{\Gamma}$ represents the actuation in the joint-space. $\bm \lambda_i $ and $\bm{J}_i$ represent the external force applied to the system and corresponding Jacobian matrix. }

\rev{Next, we choose to simplify the dynamics of the humanoid robot when interacting with the object as the full dynamics equation (\ref{eq:fullDynamics}) is highly complex and nonlinear.}
The dynamics model we developed is a simplified rigid-body dynamics (SRBD) model, shown in Figure. \ref{fig:model2}.
\rev{We consider the robot trunk, shoulders, and hips as a combined rigid body and assume to ignore the dynamical effects of the lightweight and low-inertia arms and legs \cite{di2018dynamic,li2021force}. The effects of ground reaction forces and external forces applied to the system can still be represented in this simplified model.} The control inputs $\bm u = [\bm u_h; \bm u_f]$ include pushing reaction forces $\bm u_h=[\bm F_{h,1};\:\bm F_{h,2}]$ at hands and ground reaction forces and moments $ \bm u_f=[\bm F_{f,1};\:\bm F_{f,2};\:\bm M_{f,1};\:\bm M_{f,2}]$ at the foot locations, which are applied to the float base dynamics, where $\bm F_{h,n} = [F_{h,n,x};\:F_{h,n,y};\:F_{h,n,z}], \: \bm F_{f,m} = [F_{f,m,x};\:F_{f,m,y};\:F_{f,m,z}],$ and $\bm M_{f,m} = [M_{f,m,y};\:F_{f,m,z}]$. The 5-D ground reaction force control input is previously employed in \cite{li2021force} for bipedal locomotion.

Hence, the following equations of motion 
 with respect to the robot CoM $\bm p_c$ are formed:

%%%%%%%% robot Dynamics %%%%%%%%%%%
\subsubsection{\textbf{Robot Dynamics}}
\label{subsec:dynamics}
\begin{align}
\label{eq:robotDynamicsForce}
    m_b (\ddot{\bm p}_c + \bm g)= \mathbf{\Sigma}_{n = 1}^{2}{\bm F_{h,n}} + \mathbf{\Sigma}_{m = 1}^{2}{\bm F_{f,m}}
\end{align}
\begin{multline}
    \label{eq:robotDynamicsMoment}
    \prescript{}{\mathcal{G}}{\bm I_c}\bm \dot {\bm \omega}_c = \mathbf{\Sigma}_{n = 1}^{2}( {\bm r_{h,n}^c \times \bm F_{h,n}}) 
    + \mathbf{\Sigma}_{m = 1}^{2}( {\bm r_{f,m}^c \times \bm F_{f,m}} )\\
    + \mathbf{\Sigma}_{m = 1}^{2}({\mathbf{L} \bm M_{f,m}})
\end{multline}

In equation (\ref{eq:robotDynamicsForce}), $m_b$ denotes the robot mass and $\bm g = [0;\:0;\:-g]$ is the gravity vector.
In equation (\ref{eq:robotDynamicsMoment}), $\prescript{}{\mathcal{G}}{\bm I_c}$ is the robot body moment of inertia in the world frame, $\bm r_{h,n}^c \in \mathbb{R}^3$ is the position vector from robot CoM location to $n^{th}$ hand location $\bm p_{h,n}$ in the world frame,  $\bm r_{f,m}^c \in \mathbb{R}^3$ is the position vector from robot CoM location to $m^{th}$ foot location $\bm p_{f,m}$ in the world frame, and $\mathbf L = [0, 0; 1, 0; 0, 1]$ denotes the moment selection matrix to enforce 2-D moment control input \cite{li2021force}. 

\begin{figure}[!t]
\vspace{0.2cm}
     \centering
     \begin{subfigure}[b]{0.23\textwidth}
         \centering
         \includegraphics[clip, trim=2cm 1cm 17cm 2cm, width=\columnwidth]{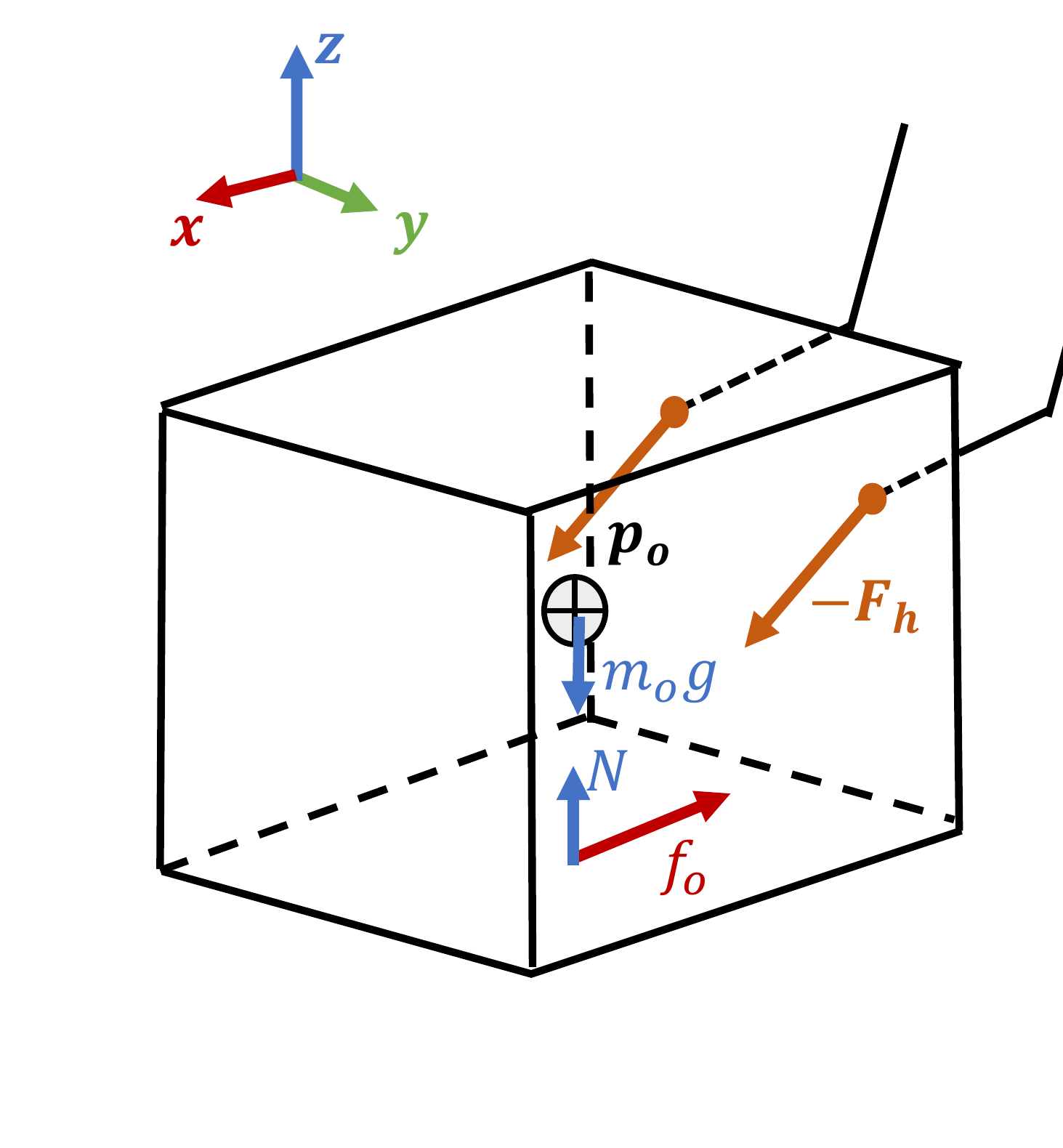}
         \caption{Object Dynamics Model}
         \label{fig:model1}
     \end{subfigure}
     \hfill
     \begin{subfigure}[b]{0.23\textwidth}
         \centering
         \includegraphics[clip, trim=18cm 1cm 2cm 2cm, width=\columnwidth]{Figures_new/Dynamics.pdf}
         \caption{Humanoid Dynamics Model}
         \label{fig:model2}
     \end{subfigure}
        \caption{{\bfseries Dynamics Model Illustrations.} a) Object under pushing, b) Humanoid SRBD}
        \label{fig:simplifiedDynamics}
        \vspace{-0.3cm}
\end{figure}

The object dynamics is also a rigid body dynamics of the object, shown in Figure. \ref{fig:model1}. The robot pushing forces, $-\bm u_h$, normal force, $\bm N$, and simplified single-point friction force, $\bm f_o \in \mathbb{R}^3$, are applied to the object,

%%%%%%%%%object dynamics%%%%%%%%%
\subsubsection{\textbf{Object Dynamics}}
\begin{align}
\label{eq:objectDynamicsForce}
    m_o (\ddot{\bm p}_o + \bm g) = -\mathbf{\Sigma}_{n = 1}^{2}{\bm F_{h,n}} - \bm R_{o,z} \bm f_o  + \bm{N}
\end{align}
\begin{align}
    \label{eq:objectDynamicsMoment}
    \prescript{}{\mathcal{G}}{\bm I_o}\bm \dot {\bm \omega}_o = \mathbf{\Sigma}_{n = 1}^{2}(\bm r_{h,n}^o \times - \bm F_{h,n}) - \bm R_{o,z}\frac{h \bm f_o}{2}
\end{align}

In equation (\ref{eq:objectDynamicsForce}), $m_o$ denotes the object mass, $\bm {N} \in \mathbb{R}^3$ stands for the normal force supporting the object. \rev{It is assumed that the friction force is approximated by single contact friction at the projection of the object's CoM on the ground and is always along the x-direction in the object frame during pushing, \rev{$\bm f_o = [\mu_g m_o g;0;0]$}.} $\mu_g$ is the coefficient of friction between the object and the ground and $\bm R_{o,z}$ is the 3-D rotation matrix defining the object's yaw angle (i.e., assuming object pitch and roll are zero during pushing). In equation (\ref{eq:objectDynamicsMoment}), $\prescript{}{\mathcal{G}}{\bm I_o}$ is the moment of inertia of the object estimated in the world frame, $\bm r_{h,n}^o \in \mathbb{R}^3$ stands for the position vector from object CoM location to $n^{th}$ hand location, and $h$ stands for the object height.

For pushing on flat ground, the object is assumed to have negligible z-direction acceleration. %Hence, equation (\ref{eq:objectDynamicsForce}), in this case, indicates $\bm N = [0; 0; -m_o g+F_{h,1,z}+F_{h,2,z}]$.

\begin{figure}[t]
\vspace{0.2cm}
\captionsetup[subfigure]{justification=centering}
     \centering
     \begin{subfigure}[b]{0.12\textwidth}
         \centering
         \includegraphics[clip, trim=0cm 4cm 24cm 3cm, width=\columnwidth]{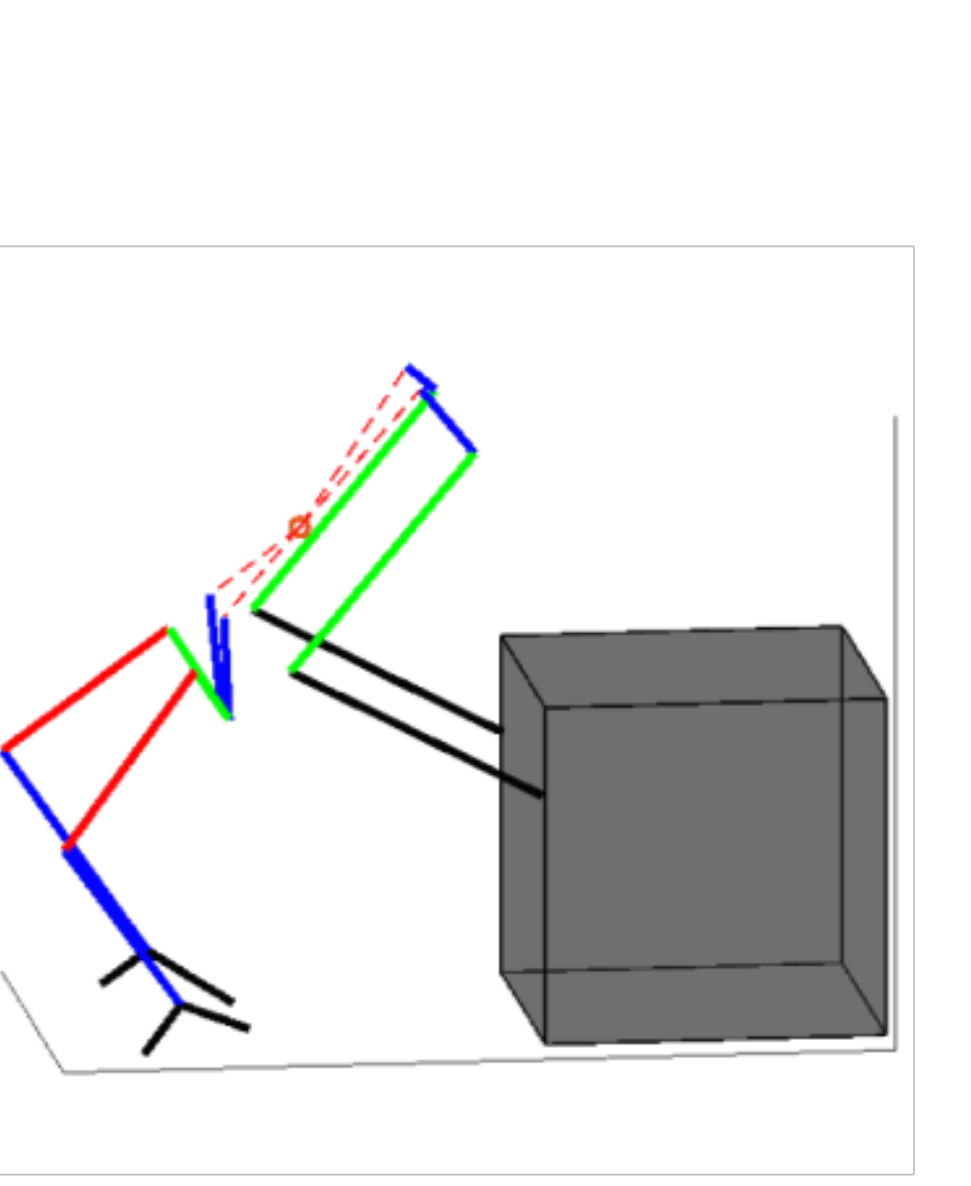}
         \caption{5 kg, 0.3 m cube, \\ $\mu_g$ = 0.7}
         \label{fig:APO1}
     \end{subfigure}
     \hfill
     \begin{subfigure}[b]{0.15\textwidth}
         \centering
         \includegraphics[clip, trim=10cm 4cm 13cm 3cm, width=\columnwidth]{Figures_new/APO3.pdf}
         \caption{10 kg, 0.5 m cube, \\ $\mu_g$ = 0.4}
         \label{fig:APO2}
     \end{subfigure}
     \hfill
     \begin{subfigure}[b]{0.20\textwidth}
         \centering
         \includegraphics[clip, trim=20.5cm 4cm 0cm 3cm, width=\columnwidth]{Figures_new/APO3.pdf}
         \caption{15 kg, 0.7 m cube, \\ $\mu_g$ = 0.5}
         \label{fig:APO3}
     \end{subfigure}
        \caption{{\bfseries Optimal humanoid Pushing Poses Solutions from Pose Optimization.} }
        \label{fig:APO}
        \vspace{-0.3cm}
\end{figure}

%%%%%%%%%% APO %%%%%%%%%%%%%%%%%%%%%
\subsection{Kinodynamics-based Pose Optimization}
\label{subsec:APO}

In this section, we will present the details of the kinodynamics-based pose optimization. 

% The reason behind utilizing whole-body flexibility and pose for pushing tasks stems from observing how humans push large and heavy objects. To generate more lateral forces against object friction, humans lean their bodies forward, lower their center of mass, and position their feet far behind their hips. In the standard pushing pose for humanoid robots (i.e., nominal standing configuration), the feet are vertically aligned with the center of mass. However, the generated moment from the reaction forces during pushing tasks can negatively affect balancing and locomotion performance, especially with heavy objects. By shifting the center of mass forward and placing the feet further back, the gravitational force from the robot body and the ground reaction forces counter the moment, providing more stable loco-manipulation.

% We use a kinodynamics-model in pose optimization, where the object-robot simplified dynamics (\ref{eq:robotDynamicsForce}-\ref{eq:objectDynamicsMoment}) in steady-state are used as well as kinematic constraints are applied.
% Pose optimization is not a discretization-based trajectory optimization method. Instead, it generates a single pose that is specific to a particular task. For simplicity and flexibility with different gait types, we have chosen to solve for humanoid pushing poses with double-leg support. Moreover, in pose optimization, we do not pass any ground reaction force and moment solutions to the MPC. Hence, this allows us to generalize the leg contact mode.

The utilization of whole-body flexibility and pose for pushing tasks is inspired by human observations. Humans lean their bodies forward, lower their center of mass, and position their feet further back to generate more force against object friction when pushing large and heavy objects. In the standard pushing pose for humanoid robots (i.e., nominal standing configuration), the feet align vertically with the center of mass. However, this configuration can negatively impact balancing and locomotion performance due to the generated moment from reaction forces during pushing, particularly with heavy objects. By shifting the center of mass forward and placing the feet further back, the gravitational force from the robot body and the ground reaction forces counteract the moment, providing more stable loco-manipulation.

In our approach, we employ a kinodynamics model in pose optimization. This model considers the simplified steady-state dynamics ((\ref{eq:robotDynamicsForce}-\ref{eq:objectDynamicsMoment})) of the object-robot system and applies kinematic constraints. Pose optimization differs from discretization-based trajectory optimization methods as it generates a single pose specific to the task at hand. To accommodate different gait types and ensure simplicity and flexibility, we focus on solving humanoid pushing poses with double-leg support. Furthermore, in pose optimization, we do not transmit ground reaction force and moment solutions to the MPC. This design choice allows for the generalization of the leg contact mode.
%%%%%%%%%%%%%%%%%

\rev{Another important feature of using the steady-state dynamics for the object model is its implication of preventing the robot from toppling over the box. The optimal pushing forces intend to prevent introducing a positive sum of moments in the y-direction with respect to the object CoM. Additionally, We choose to use the kinetic friction coefficients in equation (\ref{eq:objectDynamicsForce}-\ref{eq:objectDynamicsMoment}) to optimize for more conservative reference pushing forces, as the static friction during the initial push is generally higher than kinetic friction. Thus the resulting more conservative pushing force will prevent introducing a large moment to top over the box during the initial push. }
% Pose optimization  is a pose generator that only generates a single pose for a specific task, and it is not a discretization-based trajectory optimization. Hence we decide to solve for humanoid pushing poses with double-leg support for simplicity and versatility with different gait types. In addition, we don't pass any ground reaction force and moment solutions from pose optimization to MPC, so the leg contact mode in pose optimization is generalized. In this problem, the object shape is assumed to be rectangular. 

The pose optimization problem can be represented and solved as a NLP problem. The NLP problem's optimization variable $\mathbf X$ contains robot CoM location $\bm p_c$, body Euler angles $\bm \Theta$, joint positions $\bm q$, and control inputs $\bm u$ (i.e., pushing reaction forces $\bm F_{h,n}$, and ground reaction forces and moments $\bm F_{f,m}, \bm M_{f,m}$). The formal problem is defined as,
\begin{align}
\label{eq:cost}
\underset{\mathbf X }{\operatorname{min}} \:\:  \alpha(p_{c,z}^{ref} - p_{c,z})^2 + || \bm \Theta ||^2 _{\bm \beta} + || \bm u ||^2 _{\bm \gamma}
\end{align}

\begin{subequations}
\setlength\abovedisplayskip{-8pt}
\label{eq:constraints}
\begin{align}
\label{eq:cons1}
\operatorname{s.t.} \: \text{Steady-state Dynamics Equations (\ref{eq:robotDynamicsForce}-\ref{eq:objectDynamicsMoment}}) \\
\label{eq:cons2}
\bm q_{min} \leq \bm q \leq \bm q_{max} \quad \quad \quad \: \: \\
\label{eq:cons3}
\text{ Shoulder Location: } p_{s,n,x} \leq p_{h,n,x} \quad \: \: \: \\
\label{eq:cons4}
\nonumber 
{{ \text{Contact Friction Cone Constraints:}}} \quad \: \: \:  \\
\sqrt{ F_{h,n,y}^2 +  F_{h,n,z}^2}\leq \mu_f |F_{h,n,x}|  \quad \quad \quad \\
\label{eq:cons5}
\sqrt{ F_{f,m,x}^2 +  F_{f,m,y}^2}\leq \mu_h |F_{f,m,z}| \quad \quad \: \:\\
\nonumber 
{ \text{Pushing Contact Location Constraints:}} \\
 \label{eq:cons6}
 p_{h,n,x} = p_{o,x} - \frac{l}{2}  \quad \quad \quad  \quad \quad \quad  \quad \quad \: \:\\
 \label{eq:cons7}
p_{o,y} - \frac{w}{2} \leq p_{h,n,y} \leq p_{o,y} + \frac{w}{2} \quad \quad \quad \\
 \label{eq:cons8}
 0 \leq p_{h,n,z} \leq p_{o,z} + \frac{h}{2}  \quad \quad \quad  \quad \quad \quad \: \: \\
  \label{eq:cons9}
  p_{f,m,x} \leq p_{hip,m,x}   \quad \quad \quad  \quad \quad \quad  \quad \quad \quad\\
   \label{eq:cons10}
  p_{f,m,z} = 0   \quad \quad \quad  \quad \quad \quad  \quad \quad \quad  \quad \quad \:
 \end{align}
\end{subequations}

In equation (\ref{eq:cost}), the objectives are driving pose CoM height close to reference (i.e., standing height), minimizing body rotations, specifically for roll and yaw, and minimizing control inputs. These objectives are weighted by scalar $\alpha \in \mathbb{R}$ and diagonal matrices $\bm \beta \in \mathbb{R}^{3\times3}, \bm \gamma \in \mathbb{R}^{16\times16}$. 
\rev {In equation (\ref{eq:constraints}), $n^{th}$ hand location $\bm p_{h,n}$, shoulder location $\bm p_{s,n}$, $m^{th}$ foot location $\bm p_{f,m}$, and hip location $\bm p_{hip,m}$ are derived by forward kinematics based on joint positions.}
Equation (\ref{eq:cons2}) describes the constraint for joint angle limits. Equation (\ref{eq:cons3}) guarantees the shoulder location in the robot frame x-direction is always behind the hand location to avoid shoulder collision with the object in the optimal pose. Equation (\ref{eq:cons4}) and (\ref{eq:cons5}) ensures the contact friction constraints are satisfied. $\mu_f$ and $\mu_h$ are the dynamic friction coefficients at the foot and hand contact points. Equation (\ref{eq:cons6}) to (\ref{eq:cons8}) ensures the pushing contact locations are on the object surface facing the robot.
Equation (\ref{eq:cons9}) ensures the foot contact locations are behind the hip locations in x-direction $p_{hip,m,x}$ for an adequate stability region. Lastly, equation (\ref{eq:cons10}) constrains the foot contact locations both on the ground, as assumed double-leg contact mode in pose optimization.

%%%%%%%%%% MPC %%%%%%%%%%%%%%%%%%%%%
\subsection{Humanoid Loco-manipulation MPC}
\label{subsec:MPC}

This section presents the details of the proposed humanoid loco-manipulation MPC. 

We choose to develop a loco-manipulation MPC for the humanoid pushing problem because the locomotion MPC we introduced in \cite{li2021force} lacks the consideration of pushing reaction forces or any information from the object. With small and light objects, the locomotion MPC and manually-fixed hand location are capable of handling them as external disturbances. However, with heavy objects, the locomotion MPC cannot adapt or perform well due to the pushing reaction force being too large to be considered perturbations. 

With the loco-manipulation MPC, we intend to allow the humanoid robot to coordinate the motions of the entire body in a unified framework. We use this MPC to control manipulation and locomotion altogether to achieve pushing large, heavy, and ungraspable objects. To do so, first, we inform the MPC of the object state and directly control the object as an MPC objective, 
 secondly, the MPC is employed with the unified object-robot dynamics introduced in equation (\ref{eq:robotDynamicsForce}-\ref{eq:objectDynamicsMoment}). In addition, it optimizes the pushing and ground reaction forces while tracking the optimal pose from pose optimization. 

% The state variables include the states of both the robot and the object in the world frame $[{\bm \Theta_c};{\bm p}_c;{\bm \omega_c};\dot {{\bm p}}_c; {\bm \Theta_o};{\bm p}_o;{\bm \omega_o};\dot {{\bm p}}_o]$.

\rev{Now we present the state-space formulation of the proposed dynamics equations of the robot-object system in this MPC. We choose the state variables as a combination of robot states, object states, and gravity vector, $\bm x = [{\bm \Theta_c};{\bm p}_c;{\bm \omega_c};\dot {{\bm p}}_c; {\bm \Theta_o};{\bm p}_o;{\bm \omega_o};\dot {{\bm p}}_o; \bm g] \in \mathbb{R}^{27}$.} The state-space equation then can be written in the form of $\bm {\dot x} = \bm A \bm x + \bm B \bm u$, where
\begin{align}
\label{eq:A}
\bm A =\begin{bmatrix}
 \bm A_{c} & \mathbf{0}_{12} & \vline \bm A_{c}^{g}\\
\mathbf{0}_{12} & \bm A_{o} & \vline \bm A_{o}^{g}\\
\hline
\mathbf{0}_{3\times12} & \mathbf{0}_{3\times12}  & \vline \:\: \mathbf{0}_{3}
\end{bmatrix} \ ,
\end{align}
%%%%%%%%%%%
\begin{align}
\label{eq:Ac}
\setlength\arraycolsep{2pt}
\bm A_c = \left[\begin{array}{cccc} 
\mathbf 0_3 & \mathbf 0_3 & \mathbf S^{-1}(\bm \Theta_c) & \mathbf 0_{3} \\
\mathbf 0_3 & \mathbf 0_3 & \mathbf 0_3 & \mathbf I_3\\
\mathbf 0_3 & \mathbf 0_3 & \mathbf 0_3 & \mathbf 0_3\\
\mathbf 0_3 & \mathbf 0_3 & \mathbf 0_3 & \mathbf 0_3 \end{array} \right], 
% \mathbf S_b(\bm \Theta_c) = \left[\begin{array}{ccc}
% {c_\theta}{c_\psi} & -{s_\psi} & 0 \\
% {c_\theta}{s_\psi} & c_\psi & 0 \\
% -s_\theta & 0 & 1  \end{array} \right] 
\end{align}
%%%%%%%%%%%%%
\begin{align}
\label{eq:Ao}
\setlength\arraycolsep{2pt}
\bm A_o = \left[\begin{array}{cccc} 
\mathbf 0_3 & \mathbf 0_3 & \mathbf S^{-1}(\bm \Theta_o) & \mathbf 0_3 \\
\mathbf 0_3 & \mathbf 0_3 & \mathbf 0_3 & \mathbf I_3\\
\mathbf 0_3 & \mathbf 0_3 & \mathbf 0_3 & \mathbf 0_3\\
\mathbf 0_3 & \mathbf 0_3 & \mathbf 0_3 & \mathbf 0_3 \end{array} \right], \:
\mathbf S = \left[\begin{array}{ccc}
{c_\theta}{c_\psi} & -{s_\psi} & 0 \\
{c_\theta}{s_\psi} & c_\psi & 0 \\
-s_\theta & 0 & 1  \end{array} \right] 
\end{align}
%%%%%%%%%%%%%%%%
\begin{align}
\label{eq:Agc}
\setlength\arraycolsep{2pt}
\bm A_c^g = \left[\begin{array}{cccc} 
\mathbf 0_3, & \mathbf 0_3, & \mathbf 0_3, & \mathbf I_3 
\end{array} \right]^\intercal, 
\end{align}
%%%%%%%%%%%%%%%
\begin{align}
\label{eq:Ago}
\setlength\arraycolsep{1.5pt}
\bm A_o^g = \left[\begin{array}{cccc} 
\mathbf 0_3, & \mathbf 0_3, &  \left[\begin{array}{ccc} 0 & 
0 & \frac{-\mu_g m_o h \bm R_{o,z}}{2\prescript{}{\mathcal{G}}{\bm I_o}} \\0 & 0 & 0\\0 & 0 & 0 \end{array} \right], &  \left[\begin{array}{ccc}0 & 0 & {-\mu_g \bm R_{o,z}}\\ 0 & 0 & 0\\0 & 0 & 0 \end{array} \right]
\end{array} \right]^\intercal
\end{align}
%%%%%%%%%%%%%%%
\begin{align}
\label{eq:B}
\setlength\arraycolsep{2pt}
\bm B = \left[\begin{array}{cccccc} 
\mathbf 0_3 & \mathbf 0_3 & \mathbf 0_3 & \mathbf 0_3 & \mathbf 0_{3\times2} & \mathbf 0_{3\times2}\\
\mathbf 0_3 & \mathbf 0_3 & \mathbf 0_3 & \mathbf 0_3 & \mathbf 0_{3\times2} & \mathbf 0_{3\times2}  \\ 
\frac{\mathcal S(\bm r_{h,1}^c)}{\prescript{}{\mathcal{G}}{\bm I_c}} & \frac{\mathcal S(\bm r_{h,2}^c)}{\prescript{}{\mathcal{G}}{\bm I_c}} & \frac{\mathcal S(\bm r_{f,1}^c)}{\prescript{}{\mathcal{G}}{\bm I_c}} & \frac{\mathcal S(\bm r_{f,2}^c)}{\prescript{}{\mathcal{G}}{\bm I_c}} & \frac{\mathbf L}{\prescript{}{\mathcal{G}}{\bm I_c}} & \frac{\mathbf L}{\prescript{}{\mathcal{G}}{\bm I_c}} \\ %%%% force %%%%
\frac{\mathbf I_3}{m_b} & \frac{\mathbf I_3}{m_b} & \frac{\mathbf I_3}{m_b} & \frac{\mathbf I_3}{m_b} & \mathbf 0_{3\times2} & \mathbf 0_{3\times2}  \\ 
\mathbf 0_3 & \mathbf 0_3 & \mathbf 0_3 & \mathbf 0_3 & \mathbf 0_{3\times2} & \mathbf 0_{3\times2} \\
\mathbf 0_3 & \mathbf 0_3 & \mathbf 0_3 & \mathbf 0_3 & \mathbf 0_{3\times2} & \mathbf 0_{3\times2}   \\ %%%%% object dyn %%%%%%
-\frac{\mathcal S(\bm r_{h,1}^o)}{\prescript{}{\mathcal{G}}{\bm I_o}} & -\frac{\mathcal S(\bm r_{h,2}^o)}{\prescript{}{\mathcal{G}}{\bm I_o}} & \mathbf 0_3 & \mathbf 0_3 & \mathbf 0_{3\times2} & \mathbf 0_{3\times2} \\
-\frac{\mathbf D}{m_b} & -\frac{\mathbf D}{m_b} & \mathbf 0_3 & \mathbf 0_3 & \mathbf 0_{3\times2} & \mathbf 0_{3\times2}   \\  \end{array}  \right]
\end{align}

\rev {In equation (\ref{eq:Ac}), matrix $\mathbf S$ denotes the mapping between the time derivative of Euler angles and body angular speed, $s$ denotes sine operator, and $c$ denotes cosine operator. Note that $\mathbf S$ is not invertible at $\theta_c = \pm90^\circ$. We intend not to allow the robot to have a pitch angle of $\pm90^\circ$ in any tasks.} In equation (\ref{eq:B}), the operator $\mathcal S$ denotes the skew-symmetric matrix transformation of corresponding position vectors. $\mathbf D$ is a diagonal matrix containing $[1,1,0]$, which serves to ensure the z-direction acceleration of the object is zero. 
% The rotational inertia of the robot in the world frame $\prescript{}{\mathcal{G}}{\bm I_c}$ is approximated from body frame, $\prescript{}{\mathcal{G}}{\bm I_c} = \bm R_b^\intercal \prescript{}{\mathcal{B}}{\bm I_c} \bm R_b$. Similarly, the rotational inertia of the object in the world frame is approximated by $\prescript{}{\mathcal{G}}{\bm I_o} = \bm R_{o,z}^\intercal \prescript{}{\mathcal{B}}{\bm I_o} \bm R_{o,z}$.

To use this linear state-space dynamics equation in MPC, we discretize it at $i^{th}$ time step with step duration $dt$,
\begin{align}
\label{eq:discreteSS}
\bm {x}[i+1] = \bm {{A}}_d[i]\bm x[i] + \bm {{B}}_d[i]\bm u[i]
\end{align}
\begin{align}
\label{eq:discreteAB}
{\bm A}_d = \mathbf I_{27} + {\bm A} dt ,\quad {\bm B}_d =  {\bm B} dt
\end{align}

%%%%%% MPC form %%%%
 The formulation of the MPC problem with finite horizon $k$ is written as follows, 
\begin{multline}
\label{eq:MPCform}
%\nonumber
\underset{\bm{x,u}}{\operatorname{min}} \: \mathbf{\Sigma}_{i = 0}^{k-1} \: \{ \: \vert \vert \bm x[i+1]-  \bm x^{ref}[i+1])\vert \vert^2 _{\bm Q_i}  +  \vert \vert \bm u_f \vert \vert^2 _ {\bm{R}_i} \\
+ \mathbf{\Sigma}_{n=1}^{2} \vert \vert \bm F_{h,n}[i] - \bm F_{h,n}^{ref}[i] \vert \vert^2 _{\bm S_i}\: \}
\end{multline}

\begin{subequations}
\setlength\abovedisplayskip{-5pt}
\begin{gather}
% \vspace{-0.5cm}
\label{eq:dynamicCons}
\:\:\operatorname{s.t.} \quad  \bm {x}[i+1] = \bm {{A}}_d[i]\bm x[i] + \bm {{B}}_d[i]\bm u[i] \quad \quad \\
\nonumber 
-\mu_f'  {F}_{f,m,z} \leq  F_{f,m,x} \leq \mu_f'  {F}_{f,m,z} \quad \\
\nonumber
-\mu_f'  {F}_{f,m,z} \leq  F_{f,m,y} \leq \mu_f'  {F}_{f,m,z} \quad \\
\nonumber
-\mu_h'  {F}_{h,n,x} \leq  F_{h,n,y} \leq \mu_h'  {F}_{h,n,x} \quad \\
-\mu_h'  {F}_{h,n,x} \leq  F_{h,n,z} \leq \mu_h'  {F}_{h,n,x} \quad 
\label{eq:frictionCons}\\
\nonumber
0<  {F}_{h,min} \leq  F_{h,n,x} \leq  {F}_{h,max} \quad \:  \\
0<  {F}_{f,min} \leq  F_{f,m,z} \leq  {F}_{f,max} \quad \: 
\label{eq:forceCons}\\
\bm \tau_{min} \leq \bm J_c^\intercal \bm u \leq \bm \tau_{max} \quad \quad \quad \quad \quad \: \:
\label{eq:tauCons}\\
\label{eq:handFroceCons}
\prescript{}{\mathcal B}{F_{h,n,y}} = 0 \quad \quad \quad \quad \quad \quad \quad \quad \quad \: \\
\label{eq:swingLegCons}
\text{for swing leg:} \:\: \bm u_{f,m} = \mathbf 0 \quad \quad \quad \quad 
\end{gather}
\end{subequations}

The objectives of the MPC problem are described in equation (\ref{eq:MPCform}), which are tracking a reference trajectory based on the command, minimizing the ground reaction forces for efficient locomotion, and minimizing the deviation of optimized hand force compared to reference hand force from pose optimization. These objectives are weighted by diagonal matrices $\bm Q_i \in \mathbb{R}^{27\times27}, \bm R_i\in \mathbb{R}^{10\times10}, \bm S_i\in \mathbb{R}^{3\times3}$ .

Equation (\ref{eq:dynamicCons}) to (\ref{eq:swingLegCons}) are constraints of the MPC problem. Equation (\ref{eq:dynamicCons}) is the dynamics constraint. \rev{Equation (\ref{eq:frictionCons}) describes the conservative inscribed friction pyramid constraint of pushing and ground reaction forces, where $\mu' = \sqrt{2}\mu/2$. }Equation (\ref{eq:forceCons}) ensures the pushing and ground reaction forces are within limits. In addition, equation (\ref{eq:tauCons}) constrains the joint torque limit of the robot by using the whole-body contact Jacobian $\bm J_c$. In equation (\ref{eq:handFroceCons}), we constrain the hand force in the y-direction in the local frame to be zero, in order to avoid hand slip on the object during pushing. Equation (\ref{eq:swingLegCons}) ensures the swing leg exerts zero ground reaction force and moment.

The translation of the proposed MPC problem into Quadratic Programming (QP) form to be efficiently solved can be found in related works (e.g., \cite{di2018dynamic}, \cite{jerez2011condensed}).

%%%%%%%%%%%%%%%%%%%%%%%%%%%%%%%%%%%%%%%%%%%%%%%%%%%%%%
\subsection{Low-level Control}
\label{subsec:lowlevel}

Since the proposed loco-manipulation MPC provides the optimal pushing reaction forces for pushing and ground reaction forces and moments for the stance leg, we choose to use Cartesian PD to control hand and swing foot locations. 

The desired swing foot location $\bm p_{f,n}^{des} \in \mathbb R^{3}$ follows a heuristic foot placement policy introduced in \cite{raibert1986legged} and takes consideration of desired CoM linear speed, $\dot{\bm p}_c^{des}$ \cite{kim2019highly}. 
\rev{In addition, we adapt the heuristic foot placement by offsetting the desired capture point based on the optimal pose foot location $\bm p_{f,m}^{opt}$ and hip location $\bm p_{c}^{opt}$ for ensuring the stability region during walking.}
\begin{align}
\label{eq:footPlacement}
\bm p_{f,m}^{des} =  \bm p_{c} + (\bm p_{f,m}^{opt} - \bm p_{c}^{opt}) + \dot{\bm p}_c \frac{\Delta t}{2} + k(\dot{\bm p}_c-\dot{\bm p}_c^{des})
\end{align}
where $\Delta t$ is the gait period, and $k$ is a scaling factor for tracking desired linear velocity.

%%%%%%%%%%%%%%%%%%%%%
\begin{figure}[!t]
\vspace{0.2cm}
     \centering
     \begin{subfigure}[b]{0.5\textwidth}
         \centering
        \includegraphics[clip, trim=0cm 0cm 15cm 0.5cm, width=0.9\columnwidth]{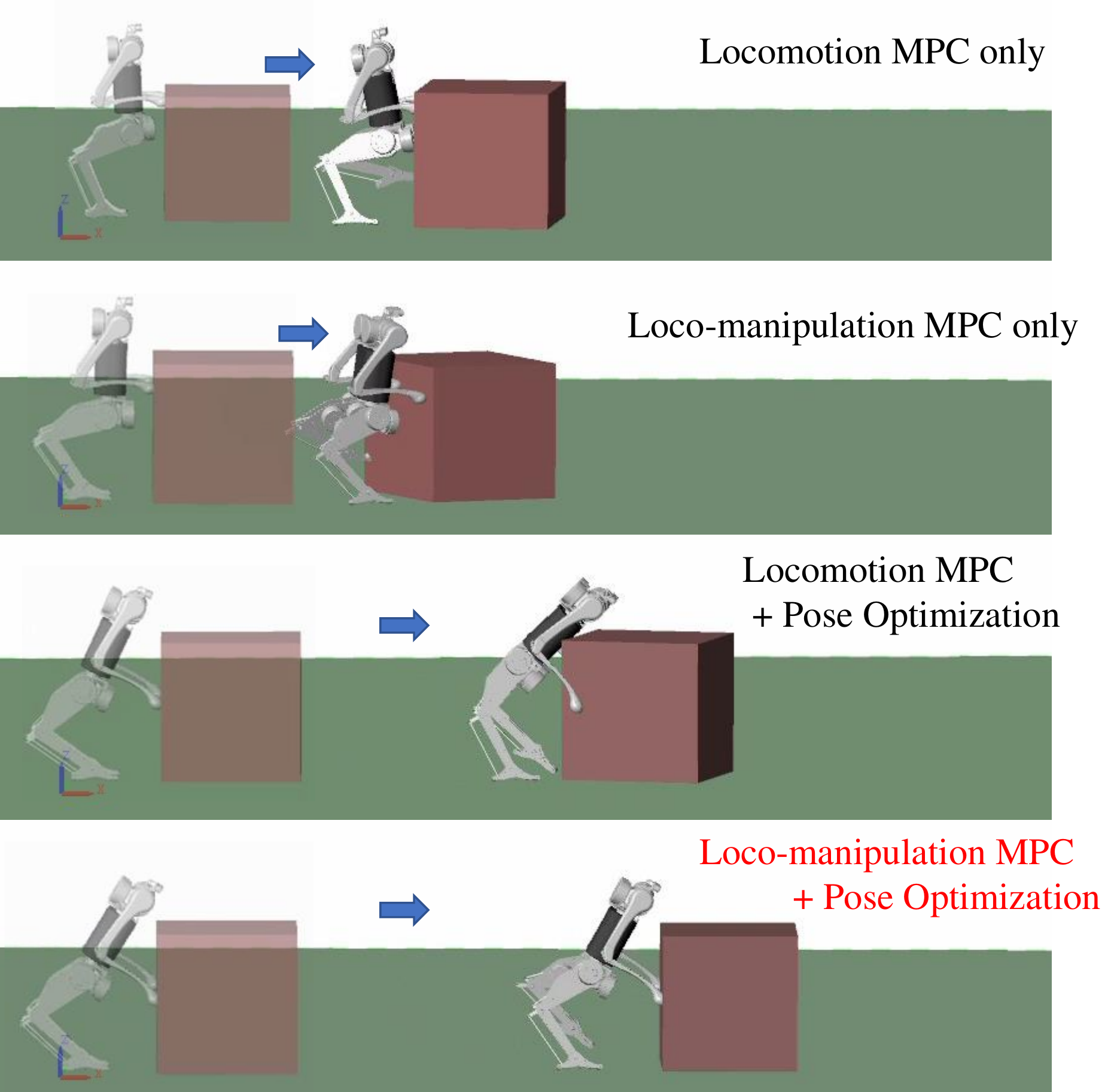}
		\caption{ Simulation snapshots. }
		\label{fig:comparison1}
     \end{subfigure}
     \hfill
     \begin{subfigure}[b]{0.45\textwidth}
         \centering
         \includegraphics[clip, trim=4cm 11cm 4cm 11cm, width=0.9\columnwidth]{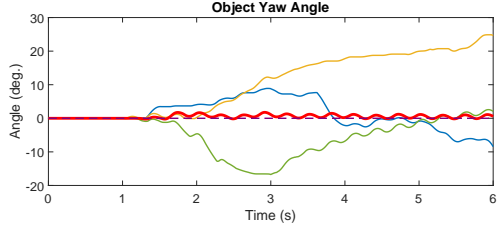}
		\caption{Object Yaw Angle Comparison Plot. }
		\label{fig:compyaw}
     \end{subfigure}
     \hfill
     \begin{subfigure}[b]{0.45\textwidth}
         \centering
         \includegraphics[clip, trim=4cm 11cm 4cm 11cm, width=0.9\columnwidth]{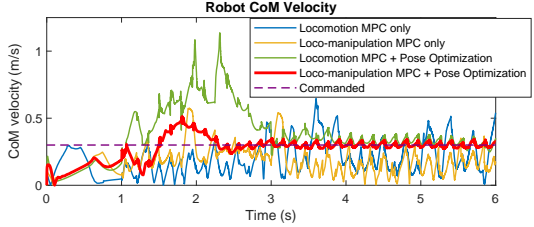}
		\caption{Robot CoM Velocity Comparison Plot. }
		\label{fig:compvx}
     \end{subfigure}
        \caption{\bfseries{Comparison of Different Approaches in 2-D Object Pushing.} }
        \label{fig:comp1}
        \vspace{-0.3cm}
\end{figure}

\begin{figure}[h]
\vspace{0.2cm}
    \center
  \includegraphics[clip,  trim=3cm 9cm 2.4cm 9.cm, width=0.95\columnwidth]{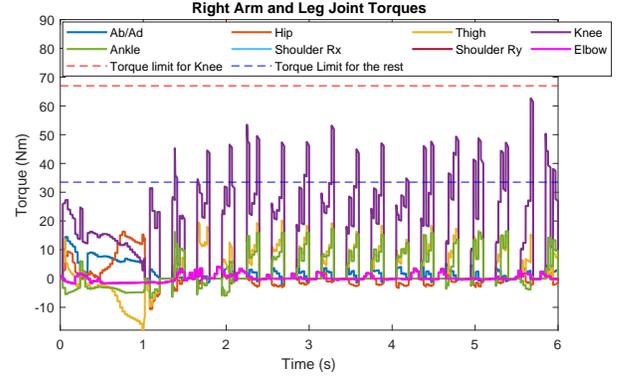}
		\caption{\bfseries{Right Arm and Leg Torque Plots in Simulation of Pushing 20 kg Object.}}
		\label{fig:20kgtorque}
		\vspace{-.2cm}
\end{figure}

%%%%%%%%%%%%%%%%%%

The Cartesian PD control law can then be applied to control $m^{th}$ swing foot to track the foot placement and $n^{th}$ hand to track optimal pushing contact location from pose optimization,
\begin{align}
\label{eq:pdlaw}
\bm F_{f,m}^{PD}=\bm K_P(\bm p_{f,m}^{des}-\bm p_{f,m})+\bm K_D(\dot{\bm p}_{f,m}^{des}-\dot{\bm p}_{f,m})
\end{align}
\begin{align}
\label{eq:pdlaw2}
\bm F_{h,n}^{PD}=\bm K_P(\bm p_{h,n}^{opt}-\bm p_{h,n})+\bm K_D(\mathbf 0-\dot{\bm p}_{h,n})
\end{align}

The aggregated PD forces $\bm u^{PD}$ is added to the optimized control inputs from MPC, 
\begin{align}
\label{eq:aggreForce1}
\bm u^* = \bm u + \bm u^{PD},
\end{align}
\begin{align}
\label{eq:aggreForce2}
\bm u_{PD} = [\bm F_{h,1}^{PD};\: \bm F_{h,2}^{PD};\: \bm F_{f,1}^{PD};\: \bm F_{f,2}^{PD};\:\mathbf{0};\:\mathbf{0}]
\end{align}

\rev{The joint torque commands are approximated by mapping the total control input $\bm u^*$ with contact Jacobians, }
\begin{align}
\label{eq:mapping}
\bm \tau = \bm J_c^\intercal \bm u^*.
\end{align}

% results
\begin{figure}[!t]
\vspace{0cm}
     \centering
     \begin{subfigure}[b]{0.5\textwidth}
         \centering
        \includegraphics[clip, trim=0cm 13cm 6cm 0cm, width=1\columnwidth]{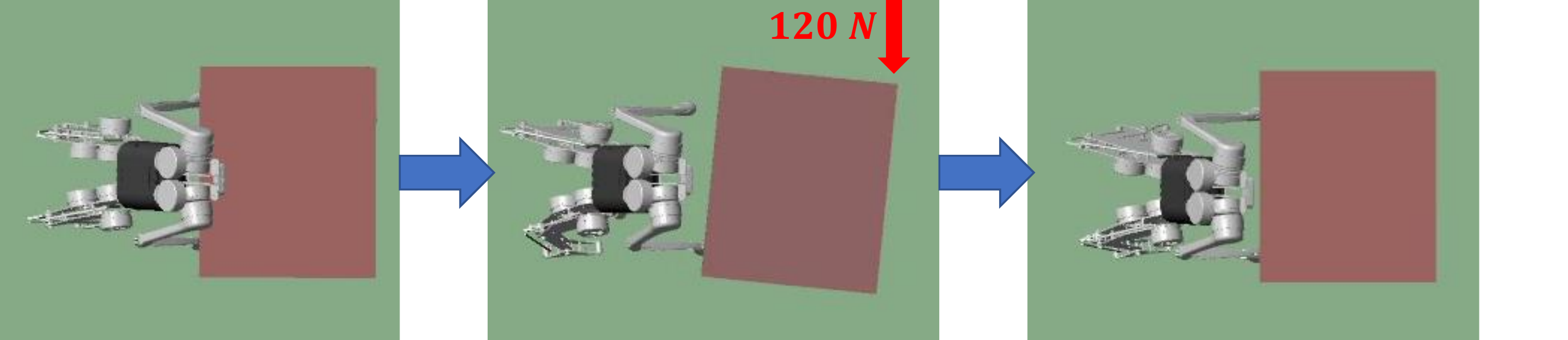}
		\caption{Simulation Snapshots. }
		\label{fig:forcesnaps}
     \end{subfigure}
     \hfill
     \begin{subfigure}[b]{0.45\textwidth}
         \centering
		\includegraphics[clip, trim=4cm 11.5cm 4cm 11cm, width=0.95\columnwidth]{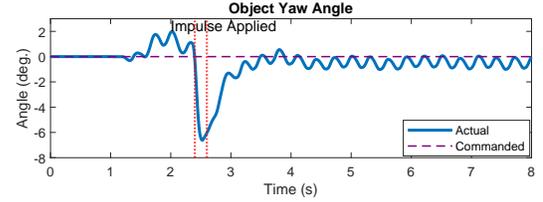}
		\caption{Object Yaw Angle Plot. }
		\label{fig:forceyaw}
     \end{subfigure}
     \hfill
     \begin{subfigure}[b]{0.45\textwidth}
         \centering
         \includegraphics[clip, trim=4cm 9.2cm 4cm 9cm, width=0.95\columnwidth]{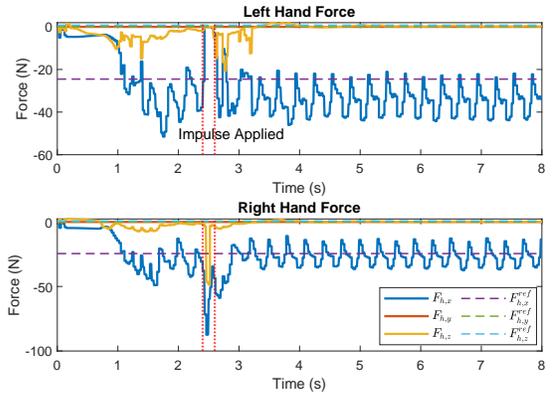}
		\caption{Hand Reaction Force Plots. }
		\label{fig:forceforce}
     \end{subfigure}
        \caption{\bfseries{Results of 2-D Pushing with External Impulse Perturbation. } }
        \label{fig:mu}
        \vspace{-0.3cm}
\end{figure}

\section{Results}
\label{sec:Results}

This section presents highlighted results of the proposed approaches in object-pushing tasks, including comparison with other approaches, pushing heavy objects, external force disturbance rejection, and tracking a desired object trajectory in 3-D.
The reader is encouraged to watch the supplementary simulation videos for better visualization of the results.

% \subsection{Setup}
To validate our proposed approaches, we set up a physical-realistic simulation framework in MATLAB Simulink with the Simscape Multibody library, \rev{which allows high-fidelity contact modeling in loco-manipulation simulations.} We use CasADi \cite{Andersson2019} to solve pose optimization as an NLP problem with \texttt{IPOPT} solver. The average solving time for pose optimization with the proposed solver, out of 50 randomized setups, is 250 $\unit{ms}$. In this simulation, we assume the state information and physical properties of the object are known.

Weighting matrices in pose optimization and MPC are,
\begin{multline}
\nonumber
    \bm Q_i = \text{diag}[1000,\:1000,\:500,\:500,\:500,\:500,\:10,\:10,\:5,\:1,\:1,\: 1,   \\
    \:0,\:0,\:1000,\: 500,\:500,\:0,\: 0,\:0,\:10,\: 1,\:1,\:0,\: 0,\: 0,\: 1],\\
    \bm R_i = \text{diag}[1,\:1,\: 1, \:1,\:1,\: 1, \:5,\:5,\: 5, \:5]\times10^{-4}, \quad \quad \quad \: \\ \bm S_i = \text{diag}[1,\:1,\: 1]\times10^{-3} \quad \quad \quad \quad \quad \quad  \quad \quad \quad \quad \quad \: \:  \:  \\
    \alpha = 50, \:\:
    \bm \beta = \text{diag}[50,\: 5, \: 50], \quad \quad \quad \quad \quad \quad \quad \quad \quad \quad  \\
    \bm \gamma = \text{diag}[1,\:1,\:1,\:1,\:1,\:1,\:1,\:1,\:1,\:1,\:1,\:1,\:5,\:5,\:5,\:5]   \:
\end{multline}

Note that the above parameters are universal to all simulation results.

% \begin{figure}[!t]
% \vspace{0.2cm}
%      \centering
%      \begin{subfigure}[b]{0.5\textwidth}
%          \centering
%         \includegraphics[clip, trim=0cm 13cm 9.5cm 0cm, width=1\columnwidth]{Figures_new/mu_snapshots.pdf}
% 		\caption{Simulation Snapshots. }
% 		\label{fig:musnaps}
%      \end{subfigure}
%      \hfill
%      \begin{subfigure}[b]{0.45\textwidth}
%          \centering
% 		\includegraphics[clip, trim=4cm 11.5cm 4cm 11.2cm, width=0.9\columnwidth]{Figures_new/mu_v3.pdf}
% 		\caption{Robot CoM Velocity Tracking Plot. }
% 		\label{fig:muv}
%      \end{subfigure}
%      \hfill
%      \begin{subfigure}[b]{0.45\textwidth}
%          \centering
%          \includegraphics[clip, trim=4cm 11.7cm 4cm 11.5cm, width=0.9\columnwidth]{Figures_new/pitch2.pdf}
% 		\caption{Robot Pitch Angle Tracking Plot. }
% 		\label{fig:pitch}
%      \end{subfigure}
%         \caption{\bfseries{Results of 2-D Pushing with Varied Ground Friction } }
%         \label{fig:mu}
%         \vspace{-0.3cm}
% \end{figure}
\begin{figure}[!t]
\vspace{0.2cm}
     \centering
     \begin{subfigure}[b]{0.4\textwidth}
         \centering
        \includegraphics[clip, trim=0cm 6cm 10cm 0cm, width=1\columnwidth]{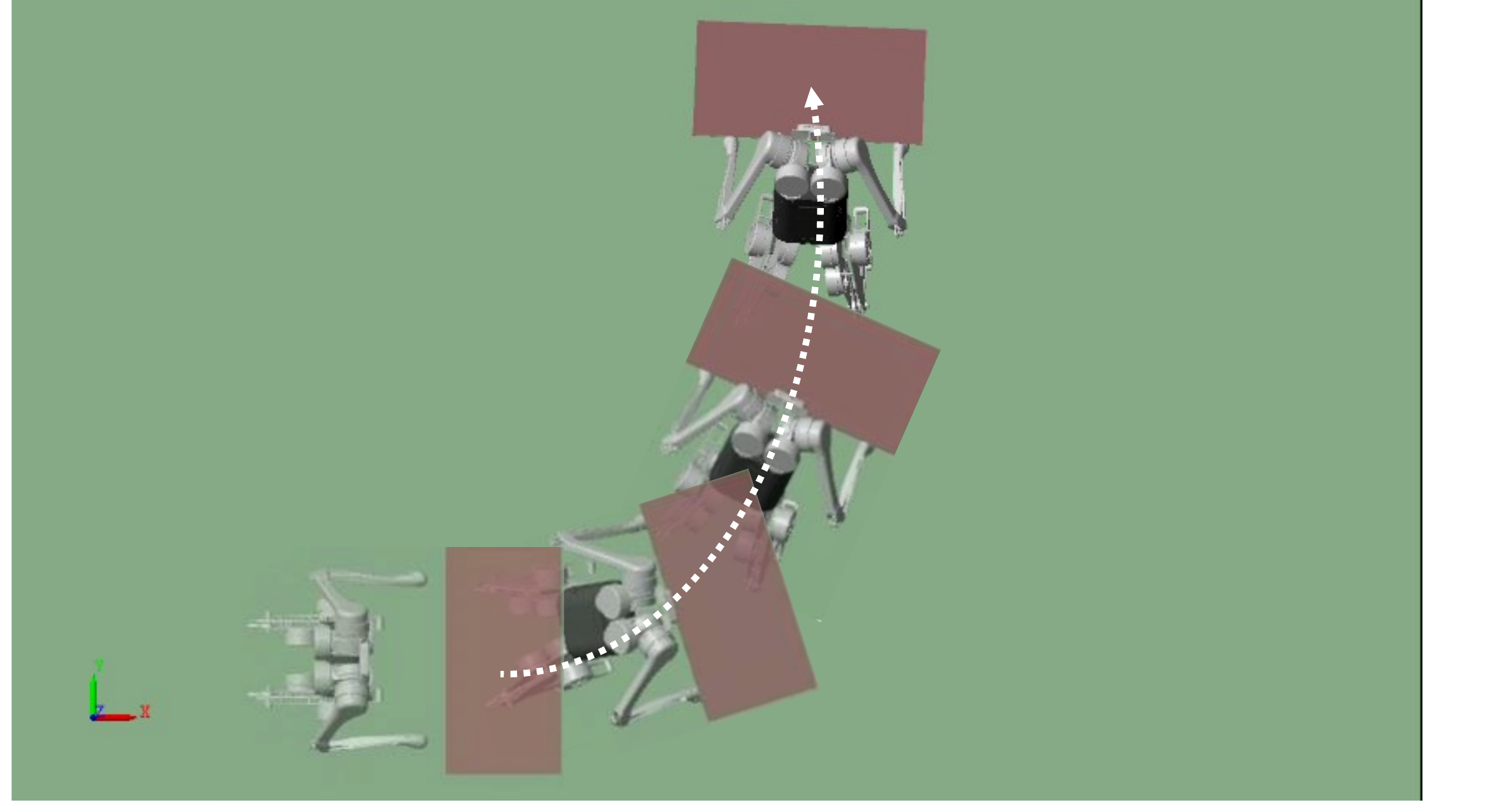}
		\caption{Simulation Snapshots. }
		\label{fig:3dsnaps}
     \end{subfigure}
     \hfill
     \begin{subfigure}[b]{0.45\textwidth}
         \centering
		\includegraphics[clip, trim=4cm 11.7cm 4cm 11.4cm, width=0.95\columnwidth]{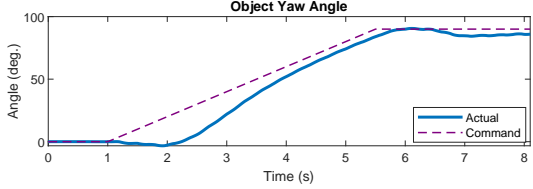}
		\caption{Object Yaw Angle Plot. }
		\label{fig:3dyaw}
     \end{subfigure}
     \hfill
     \begin{subfigure}[b]{0.45\textwidth}
         \centering
         \includegraphics[clip, trim=3cm 9.6cm 2.7cm 9.4cm, width=0.95\columnwidth]{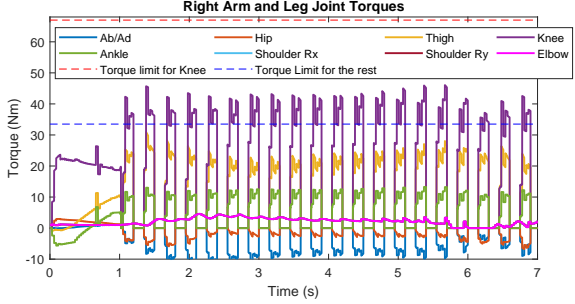}
		\caption{Right Arm and Leg Joint Torque Plot. }
		\label{fig:3dtorque}
     \end{subfigure}
        \caption{\bfseries{Results of 3-D Pushing with 90-degree Turn. } }
        \label{fig:3d}
        \vspace{-0.3cm}
\end{figure}

\subsection{Comparison of Approaches in 2-D Pushing}
%% Comparison between two models
Firstly, we present a set of comparisons between different approaches in the humanoid robot 2-D pushing problem. The approaches we compared are
\begin{enumerate}
    \item Locomotion MPC only; 
    \item Loco-manipulation MPC only;
    \item Locomotion MPC and pose optimization;
    \item Loco-manipulation MPC and pose optimization (proposed approach).
\end{enumerate}
We compare these approaches by controlling the robot to push a 10 $\unit{kg}$ box on flat ground in the x-direction, with $\mu_g = 0.5$. Figure. \ref{fig:comparison1} shows the snapshots of the simulations. It is observed that the baseline locomotion MPC in \cite{li2021force} does not perform well since it does not optimize pushing force for loco-manipulation. The loco-manipulation MPC-only approach does not have an optimal hand location for pushing, resulting in ineffective corrections to the box yaw angle deviation. Combining the pose optimization and locomotion MPC, it is observed a better velocity tracking performance, as shown in Figure. \ref{fig:compvx}. However, since hand forces and box states are not considered in the control framework, this approach struggles to keep the hands on the object during flight and failed due to the hand slipping out of the pushing surface. The proposed loco-manipulation MPC and pose optimization framework allow very stable pushing. With this proposed approach, it can be observed in Figure. \ref{fig:compyaw} and \ref{fig:compvx}, the box yaw angle stays around zero, and velocity follows the command.

% \subsection{Robustness to Ground Friction Perturbations}
% With the loco-manipulation MPC, we intend to demonstrate the robustness in overcoming terrain perturbation such as ground friction variations during the course in 2-D pushing. Figure. \ref{fig:musnaps} illustrates the simulation snapshots over the flat terrain made up of patches that have varied friction coefficients. The  tracking plots in Figure. \ref{fig:muv} and \ref{fig:pitch} show that the robot has good velocity and pitch angle tracking performance even without knowing the ground friction change during the course.

\subsection{Versatility of Pose Optimization}
Pose optimization can generate different pushing poses for a range of object parameters. Object side lengths ($l,w,h$) range from 0.3 $\unit{m}$ to 1 $\unit{m}$. Object mass $m_o$ ranges from 1 $\unit{kg}$ to 20 $\unit{kg}$. Object-ground friction coefficient $\mu_g$ ranges from 0.2 to 0.7.  Figure. \ref{fig:APO} illustrates pushing poses generated by pose optimization based on varied object parameters, and Figure. \ref{fig:title} shows their corresponding simulation snapshots. 
Our proposed approach is capable of allowing our humanoid robot to push a very large and heavy object that has a side length of 1 $\unit{m}$ and mass of 20 $\unit{kg}$ (118$\%$ robot mass). The torque plot for pushing such objects in simulation with a desired velocity of 0.3 $\unit{m/s}$ is shown in Figure. \ref{fig:20kgtorque}. It can be observed that the motor torque limits are not exceeded during this task.

\subsection{Robustness to External Disturbances}
The proposed loco-manipulation MPC is capable of compensating for unknown external impulses applied to the system and recovering from such disturbances. Additionally, this MPC considers the object trajectory, enabling it to correct for perturbations to the object through the online controller. Simulation snapshots in Figure. \ref{fig:forcesnaps} demonstrates the control framework's ability to recover from \rev{an external impulse (120 $\unit{N}$ for 0.3 $\unit{s}$)} applied to the object . In the object yaw angle plot in Figure. \ref{fig:forceyaw}, it is shown that the proposed control framework reacts to the box yaw deviation quickly and effectively. In Figure. \ref{fig:forceforce}, it can be observed that the optimized pushing reaction forces from MPC stay close to the reference value from pose optimization during flight. During the application of external impulse, the controller has the knowledge of the deviation in object yaw angle, which results in zero reaction forces on the left hand and increased force magnitude on the right hand to correct this deviation.

\subsection{3-D Loco-manipulation}
We have also extended our framework to 3-D which allows the robot to push an object given a specific trajectory or command to MPC. Figure. \ref{fig:3dsnaps} shows the simulation snapshots of 3-D loco-manipulation with a 90-degree turn while pushing a 10 $\unit{kg}$ object with $\mu_g = 0.4$. Figure. \ref{fig:3dyaw} presents the object yaw angle tracking with the proposed MPC. Lastly, Figure. \ref{fig:3dtorque} shows that the joint torque commands in the left arm and left leg are within limits.

% Conclusion
%!TEX root = ../Main.tex

\section{Conclusions}
\label{sec:Conclusion}

% In conclusion, we introduce a kinodynamics-based pose optimization and loco-manipulation MPC framework on humanoid robots to tackle the problem of pushing heavy and large objects by leveraging humanoid whole-body poses and synchronized locomotion and manipulation control. The pose optimization framework is developed to consider the kinodynamics constraints of the object-robot system and solve for an optimal pushing pose and reference pushing forces. We then pair it with loco-manipulation MPC to coordinate pushing and walking motions by optimizing both pushing reaction forces and ground reaction forces. 
% The proposed method has been validated numerically in simulations.
% The proposed control framework allows the humanoid robot to push differently sized and weighted objects with different poses. The control framework has shown robustness in loco-manipulation when external force perturbations are introduced.
% In addition, we have demonstrated effectiveness in tracking commanded state targets in 3-D and correcting state deviations in various simulations. 

In conclusion, we propose a kinodynamics-based pose optimization and loco-manipulation MPC framework for humanoid robots to push heavy and large objects. The framework combines whole-body poses, coordinated locomotion and manipulation control. The offline pose optimization considers kinodynamics constraints to find an optimal pushing pose and reference forces. Then the online loco-manipulation MPC optimizes pushing and ground reaction forces for real-time control. Numerical simulations validate the framework's effectiveness in pushing objects of different sizes and weights, handling external force perturbations, tracking state targets, and correcting state deviations.

% \input{Sections/Acknowledgement}
% \newpage
\balance
\bibliographystyle{ieeetr}
\bibliography{reference.bib}

% Document end
\end{document}